\documentclass[letterpaper]{article} 
\usepackage{aaai25}  
\usepackage{times}  
\usepackage{helvet}  
\usepackage{courier}  
\usepackage[hyphens]{url}  
\usepackage{graphicx} 
\usepackage{natbib}  
\usepackage{caption} 
\usepackage{booktabs}
\usepackage{amsmath}
\usepackage{multirow}
\usepackage{algorithm}
\usepackage{algorithmic}
\usepackage{amssymb} 
\usepackage{pifont} 
\usepackage{xcolor}

\usepackage{newfloat}
\usepackage{listings}
\DeclareCaptionStyle{ruled}{labelfont=normalfont,labelsep=colon,strut=off} 
\floatstyle{ruled}
\newfloat{listing}{tb}{lst}{}
\floatname{listing}{Listing}
\usepackage{bibentry}
\begin{document}
\title{\textit{HiPrompt}: Tuning-free Higher-Resolution Generation with \underline{Hi}erarchical \\ MLLM Prompts}
\author {
    Xinyu Liu$^1$,
    Yingqing He$^1$,
    Lanqing Guo$^2$,
    Xiang Li$^3$,
    Bu Jin$^4$,
    Peng Li$^1$,
    Yan Li$^1$,
    Chi-Min Chan$^1$,
    Qifeng Chen$^1$,
    Wei Xue$^1$, Wenhan Luo$^1$, Qifeng Liu$^1$, Yike Guo$^1$
}
\affiliations {
    \textsuperscript{\rm 1}Hong Kong University of Science and Technology\\
    \textsuperscript{\rm 2}Nanyang Technological University\\
    \textsuperscript{\rm 3}Tsinghua University\\
    \textsuperscript{\rm 4}University of Chinese Academy of Sciences\\
    \tt\small Project page: \url{https://liuxinyv.github.io/HiPrompt/}
}




\twocolumn[{
    \renewcommand\twocolumn[1][]{#1}
    \begin{center}
        \centering
        \maketitle
        \includegraphics[width=1.0\textwidth]{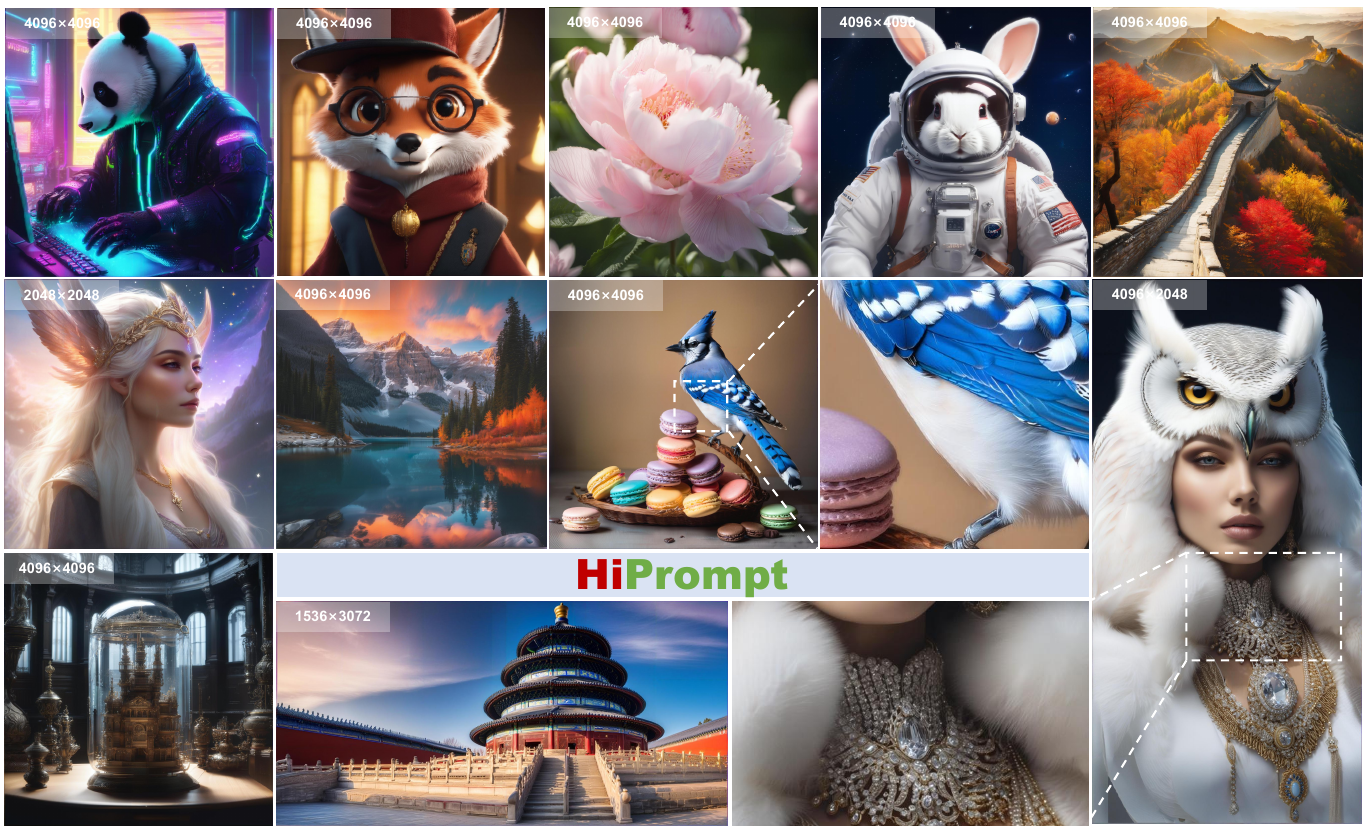}
        \captionof{figure}{Examples of HiPrompt at various higher resolutions based upon SDXL. SDXL can synthesize images up to a resolution of $1024^2$, while our method extends SDXL to generate images at $4\times$, $16\times$ without any fine-tuning. Please zoom in for a better view.}
        \label{fig:illustration}
    \end{center}
}]
\begin{abstract}

The potential for higher-resolution image generation using pretrained diffusion models is immense, yet these models often struggle with issues of object repetition and structural artifacts especially when scaling to 4K resolution and higher. 
We figure out that the problem is caused by that, a single prompt for the generation of multiple scales provides insufficient efficacy.
In response, we propose HiPrompt, a new tuning-free solution that tackles 
the above problems by introducing hierarchical prompts.
The hierarchical prompts offer both global and local guidance.
Specifically, the global guidance comes from the user input that describes the overall content, while the local guidance utilizes patch-wise descriptions from  MLLMs to elaborately guide the regional structure and texture generation.
Furthermore, during the inverse denoising process, the generated noise is decomposed into low- and high-frequency spatial components. These components are conditioned on multiple prompt levels, including detailed patch-wise descriptions and broader image-level prompts, facilitating prompt-guided denoising under hierarchical semantic guidance.
It further allows the generation to focus more on local spatial regions and ensures the generated images maintain coherent local and global semantics, structures, and textures with high definition.
Extensive experiments demonstrate that HiPrompt outperforms state-of-the-art works in higher-resolution image generation, significantly reducing object repetition and enhancing structural quality. 
\end{abstract}

%

\section{Introduction}

Stable Diffusion (SD)~\cite{rombach2022highresolutionimagesynthesislatent} has garnered widespread attention and led to wide adoption, particularly in the field of text-to-image (T2I) generation~\cite{teng2023relay,lu2024fit,zhang2023show,wang2023lavie,lu2024fit,ding2023patched,gu2023matryoshka}. At the same time, the demand for high-resolution images has surged, driven by advanced display and the need for detailed visuals.
However, to generate images at resolutions higher than the training resolution of SDXL~\cite{podell2023sdxl}, retraining the model or training a new super-resolution model is both resource-extensive and time-consuming.

Existing works ~\cite{zhang2021designing, he2023scalecrafter, bar2023multidiffusion, si2024freeu, du2024demofusion} have investigated training-free paradigms to generate higher-resolution images, aiming to address the challenge of substantial computational resource and time requirements.
A series of patch-based text-to-image-generation approaches~\cite{bar2023multidiffusion, du2024demofusion, lin2024accdiffusion} have been explored, which fuse multiple overlapping denoising paths.
However, they are plagued by pattern repetition and structure artifact problems.
For instance, MultiDiffusion~\cite{bar2023multidiffusion} imports severe object repetition 
because of the integration of the controls from all
regions into a generation process.
Although DemoFusion~\cite{du2024demofusion} attempts to keep an accurate global structure by incorporating global semantic information through residual connections and dilated sampling, it still suffers from the object repetition issue and incorrect local structures.
We figure out that the object repetition issue is caused by the unmatched semantics between the input prompt with the local patches during its patch-based denoising process: The input prompt tends to describe the overall content, whereas the patch generation at a higher scale contains only local objects.
This motivates us to propose hierarchical prompts to accurately guide the higher-resolution image generation at different scales.

\begin{figure}[h]
    \centering
    \includegraphics[width=1.\linewidth]{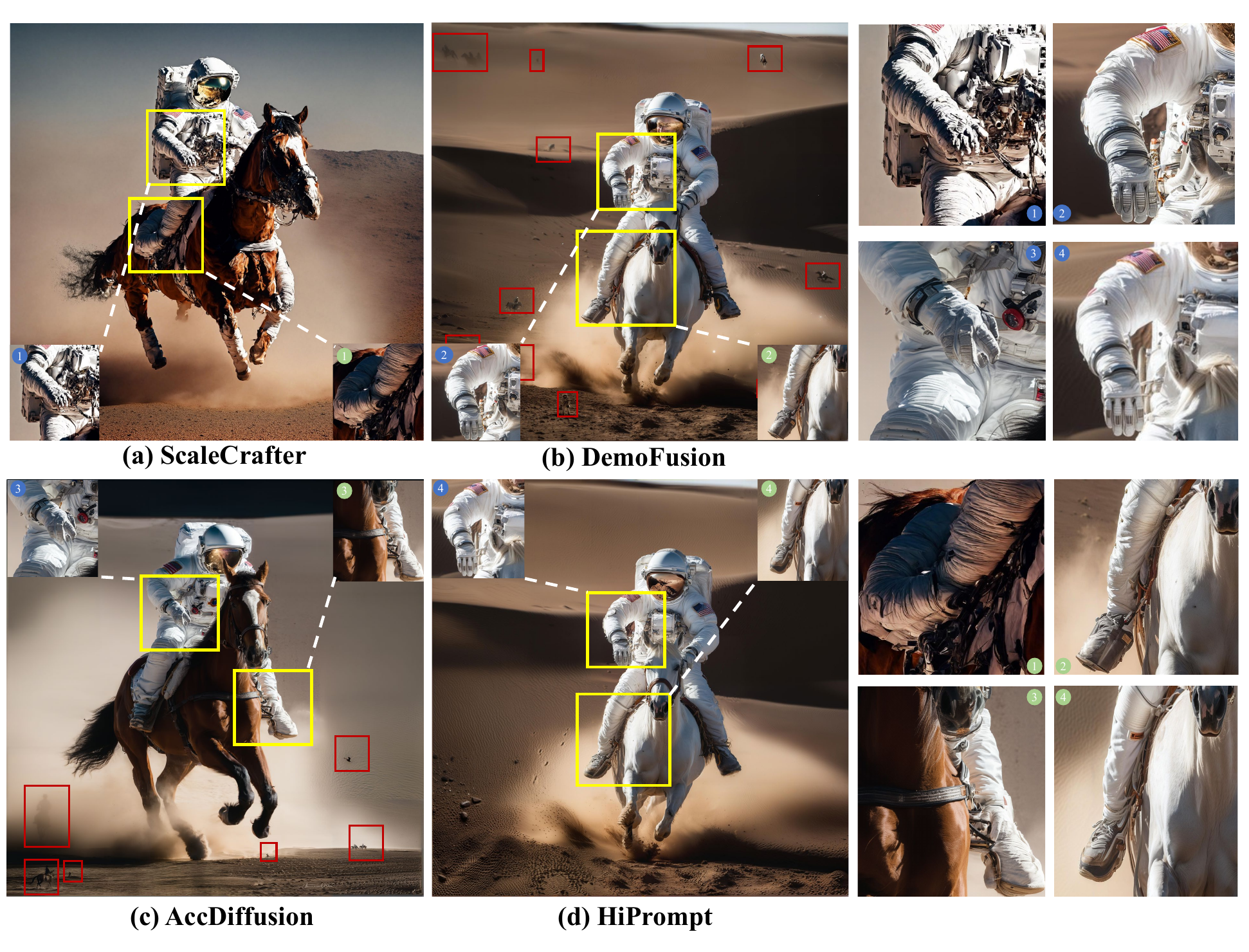}
    \vspace{-3mm}
    \caption{Visual comparison between ScaleCrafter~\cite{he2023scalecrafter}, DemoFusion~\cite{du2024demofusion}, AccDiffusion~\cite{lin2024accdiffusion}, and HiPrompt.
    Under setting of $16\times$ ($4096^2$). The red boxes highlight the repeated object problem, while the yellow boxes denote areas with blurred and unreasonable structures.}
    \label{fig:teaser}
\end{figure}

Most recently, AccDiffusion~\cite{lin2024accdiffusion} proposes patch-content-aware prompts and dilated sampling with window interaction but still suffers from small object repetition, along with blurriness and irrelevant content generation due to the loss of detailed text guidance, as depicted in Figure~\ref{fig:teaser}.
It is evident that AccDiffusion still exhibits issues with small object repetition and local blurriness. 
This is primarily due to AccDiffusion's reliance on the attention map to remove words absent from the image. However, the attention map's insufficient distinction of word responses leads to incomplete removal of repeated words and results in less detailed text descriptions. 

In this work, we introduce a hierarchical MLLM prompts-based tuning-free diffusion model, an innovative and effective approach that eradicates pattern repetition and artifacts using hierarchical prompts that offer both global and local guidance. Specifically, the global guidance comes from the user input that describes the overall content, while the local guidance utilizes patch-wise descriptions from  MLLMs to elaborately guide the local structure and texture generation. 
We adopt different MLLMs (LLAVA~\cite{liu2024improved}, ShareCaptioner~\cite{chen2023sharegpt4v}) to verify the generalization and effectiveness of HiPrompt. 

To enhance the consistency between the local and global aspects of an image, we decompose the noisy image into low and high-frequency components. This decomposition facilitates parallel denoising, with hierarchical prompts used to control each element via diffusion model sampling. Specifically, low-frequency components are conditioned on global prompts, while high-frequency components are directed by prompts generated through MLLMs. This method aims to yield high-resolution images with improved detail and structural integrity.
As shown in Figure~\ref{fig:illustration}, HiPrompt generates high-quality images at various higher resolutions and effectively resolves object repetition while preserving detailed and coherent structures even upon zooming in.

To sum up, our contributions are as follows.
\begin{itemize}
    \item We present hierarchical prompt (HiPrompt), hierarchical semantic guidance for the tuning-free higher-resolution generation. HiPrompt corrects the unmatched semantics between global prompt and local patches, thus solving the issue of object repetition of previous works.
    \item We explore the decomposition of images into spatial frequency components, conditions on fine-grained local and broad-scale global prompts, and parallel denoise.
    HiPrompt facilitates spatially controlled prompting, thereby ensuring the preservation of local-global structural and semantical coherence in higher-resolution image generation. 
    \item 
    We provide extensive quantitative and qualitative evaluations that compare HiPrompt with previous state-of-the-art methods, demonstrating the effectiveness of HiPrompt.
\end{itemize}

\section{Related Work}
\subsection{Text-to-Image Synthesis}


Text-to-image generation models~\cite{ding2021cogview,li2024playground,geng2024visual} have gained considerable prominence due to the notable advancements achieved with denoising diffusion probabilistic models~\cite{song2020denoising,ho2020denoising}. 
Recent text-guided generation models~\cite{podell2023sdxl,he2023scalecrafter,feng2024ranni,chen2024training}, based on latent diffusion models (LDMs), exhibit a remarkable ability to produce high-quality images. 
These models enhance image fidelity by iteratively refining a noisy input through denoising processes, with the generation guided by textual prompts that ensure detailed and contextually accurate results.

Building on these advances, the field of high-resolution image generation~\cite{guo2024make,zheng2024any,chen2024image,zheng2024any,xie2023difffit} has seen the emergence of several innovative approaches. For instance, Imagen~\cite{saharia2022photorealistic} and Stable Diffusion~\cite{rombach2022high} have introduced additional super-resolution networks to improve image resolution. 
In contrast, recent models like SDXL~\cite{podell2023sdxl} and PixArt-$\alpha$~\cite{chen2023pixart}
strive to directly generate high-resolution images in a single stage using end-to-end approaches. However, these models are still limited when it comes to ultra-high resolution, such as 4K, due to the enormous complexity of generation and the lack of sufficiently large-scale ultra-high resolution training data.


\subsection{Tuning-Free Higher-Resolution Generation}
The synthesis of high-resolution images presents a formidable challenge due to the intrinsic complexities of learning from high-dimensional data and the substantial computational resources necessary to extend image generation beyond the trained resolution.
Most recently, some training-free approaches~\cite{he2023scalecrafter,bar2023multidiffusion,si2024freeu,du2024demofusion,zhang2023hidiffusion,guo2024make,yang2024mastering,wang2024generative,jin2024training} adjust inference strategies or
network architectures for higher-resolution generation to add sufficient details to produce high-quality and high-resolution results.

ScaleCrafter~\cite{he2023scalecrafter} proposes a re-dilation strategy for dynamically increasing the receptive field in the diffusion UNet~\cite{ronneberger2015u}. 
The Patch-based method~\cite{bar2023multidiffusion} proposes a multi-stage diffusion process that progressively refines the generated image. DemoFusion~\cite{du2024demofusion} further improves upon this approach by introducing a progressive generation strategy that aligns residual connection and dilated sampling. 
Approaches like  ~\cite{huang2024fouriscale} introduce a fusion strategy from a frequency perspective which focuses on operations within convolutions. 
~\cite{shi2024resmastermasteringhighresolutionimage} is a concurrent work with us.
Compared to its approach, which performs structural control in the frequency domain, our method processes noise decomposition into high and low frequencies directly in the image domain. Moreover, HiPropmt conducts parallel denoising and applies a Gaussian kernel to extract the high and low frequencies corresponding to multi-scale prompts. This design allows for distinct control over structure and detail.  Additionally, to obtain accurate and dense prompt, we further filter and refine the hierarchical prompts from MLLMs.



\section{Methodology}
\begin{figure*}[h]
    \centering
    \includegraphics[width=0.95\linewidth]{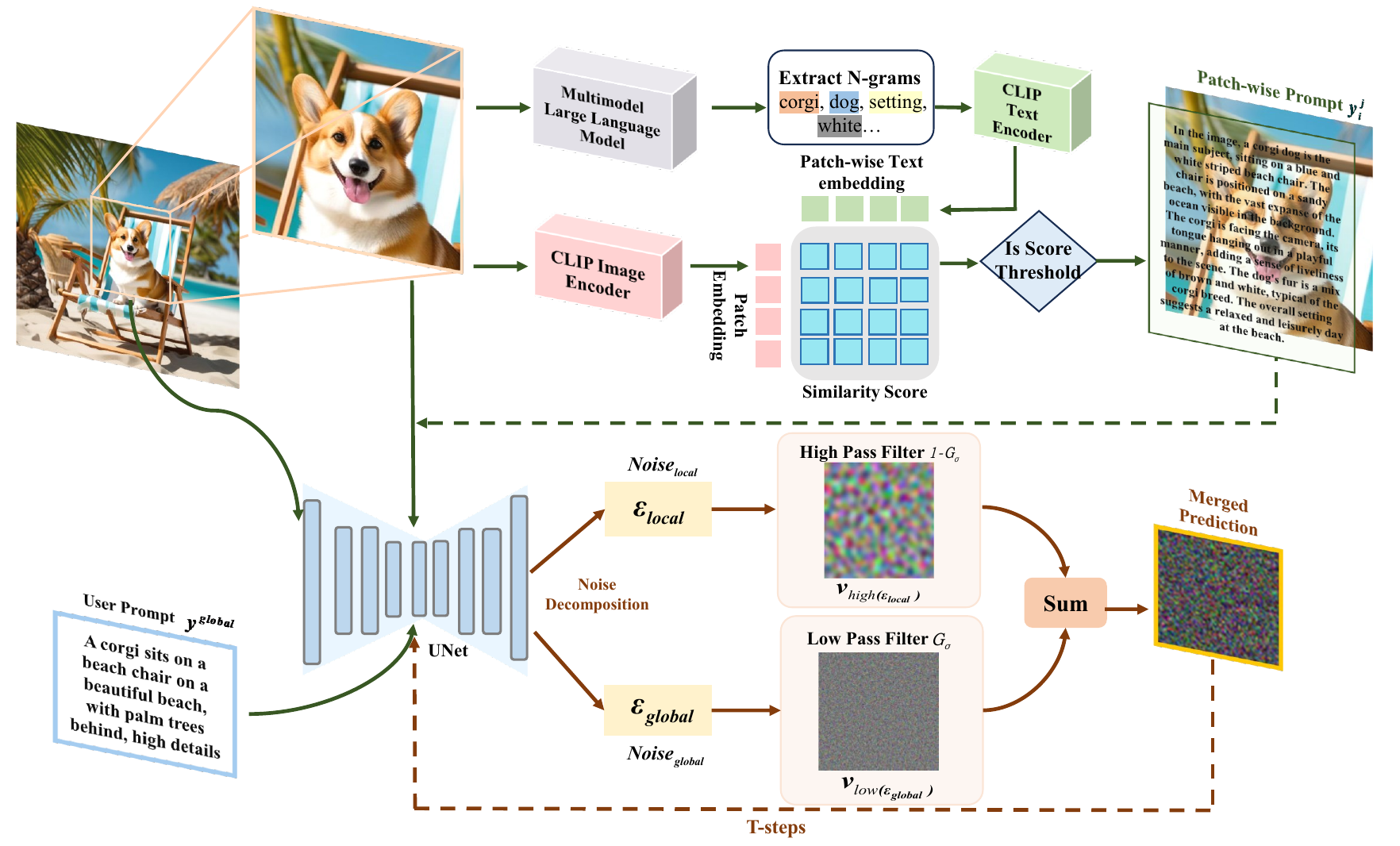}
    \vspace{-3mm}
    \caption{The overall framework of HiPrompt. The upper portion illustrates the hierarchical prompt generation module, while the lower section outlines the noise decomposition process. Given a low-resolution image, MLLMs are employed to generate dense local descriptions for each overlapping local patch. To enhance the quality of these detailed prompts, we utilize N-grams $(n = 1)$ refinement to filter out irrelevant noise. Subsequently, HiPrompt decomposes the noisy image into low- and high-spatial frequency components using low-pass and high-pass Gaussian filters. These components are denoised in parallel, conditioned on the hierarchical prompts, and then summarized into final estimation during the inverse denoising process.}
    \label{fig:architecture}
\end{figure*}
\subsection{Preliminaries}

\textbf{Latent Diffusion Model.} Diffusion models generate data by progressively refining noisy samples. Starting from Gaussian noise, the model iteratively removes noise over $T$ time steps, ultimately producing a clean sample at the final step. During this process, the noise level is controlled by a variance schedule, which dictates the amount of noise present in the sample at each intermediate time step $t$.


Following this, the two core components of the diffusion model, the diffusion and the denoising process, take place in the latent space.  With a prescribed variance schedule $\beta_1, \cdots, \beta_T$, the diffusion process can be formulated as
\begin{equation}~\label{}
    q(\mathbf{z}_t|\mathbf{z}_{t-1})=\mathcal{N}(\mathbf{z}_t;\sqrt{1-\beta_t} \mathbf{z}_{t-1}, \beta_t \mathbf{I}).
\end{equation}
In contrast, the denoising process aims to recover the cleaner version $\mathbf{z}_{t-1}$ from $\mathbf{z}_t$ by estimating the noise, which can be expressed as 
\begin{equation}~\label{}
    p_\theta(\mathbf{z}_{t-i}|\mathbf{z}_t)=\mathcal{N}(\mathbf{z}_{t-1};\mathbf{\mu}_\theta(\mathbf{z}_t, t), \mathbf{\Sigma}_\theta(\mathbf{z}_t, t)),
\end{equation}
where $\mathbf{\mu}_\theta$ and $\mathbf{\Sigma}_\theta$ are determined through estimation procedures and $\theta$ denotes the parameters of the denoise model.

\vspace{1mm}
\noindent\textbf{MultiDiffusion}~\cite{bar2023multidiffusion} 
achieves high-resolution image generation by integrating multiple overlapping denoising paths. Given a latent representation \( \mathbf{z}_t \in \mathbb{R}^{M' \times N' \times C} \) of a high-resolution image, 
MultiDiffusion~\cite{bar2023multidiffusion} employs a sliding window strategy to sample patches from \( \mathbf{z}_t \). This results in a set of patch noises \(\{ \mathbf{z}_t^{(i)} \}_{i=1}^{Q} \), where each patch \( \mathbf{z}_t^{(i)} \in \mathbb{R}^{M \times N \times C} \) and $Q$ is the total number of patches.
Patch-wise denoising is then performed to obtain
\(\{ \mathbf{z}_{t-1}^{(i)} \}_{i=1}^{Q} \). These denoised patches \(\{ \mathbf{z}_{t-1}^{(i)} \}_{i=1}^{Q} \) are subsequently recombined into \( \mathbf{z}_{t-1} \) by averaging the overlapping regions. Finally, a high-resolution image is obtained by decoding \( \mathbf{z}_0 \) into the output image \( \mathbf{x}_0 \).


\subsection{Overview}
As illustrated in Figure~\ref{fig:architecture}, we first introduce a hierarchical prompt-based diffusion model, which recaptions low-resolution image patches with dense and localized descriptions derived from MLLMs (like LLAVA~\cite{liu2024improved} and ShareCaptioner~\cite{chen2023sharegpt4v}) to mitigate repetitive patterns and increase detail accuracy. Then, given the noisy image, we propose to decompose an image into low and high spatial frequencies, corresponding to global and local prompts, to parallel denoise during the inverse denoising process. Noises controlled by hierarchical prompts are consolidated into a single, combined estimated prediction.

\subsection{Hierarchical Prompt Guidance}
Following Demofusion~\cite{du2024demofusion}, we employ SDXL~\cite{podell2023sdxlimprovinglatentdiffusion} to create a low-resolution image based on the user prompt $\mathbf{y}^\text {global}$.
The image is upsampled to the target resolution, which is then divided into \(Q\) overlapping patches.
Instead of relying solely on low-resolution images and global prompts as generation guidance, we introduce hierarchical prompts for each low-resolution image patch to provide more detailed and accurate guidance. We investigate a set of patch-wise prompts $\left\{\mathbf{y}^j\right\}_{j=0}^Q$
recaptioned from the MLLMs, e.g., LLAVA and ShareCaptioner, where $\mathbf{y}^j$ is responsible for injecting specific conditions into the corresponding image patch. 
By adopting this approach, we can generate more detailed and nuanced information for each patch, thereby enhancing the fidelity of the generated image and minimizing the semantic gap between the prompt and the final results.

During the re-caption process, the query for LLAVA follows the template: ``Here's a formula for a Stable Diffusion image prompt: an image of [adjective] [subject] [material], [color scheme], [photo location], detailed. Answer in one sentence.'' Concurrently, the instruction for ShareCaptioner is: ``Analyze the image in a comprehensive and detailed manner.'' Following this instruction, the MLLMs will generate a more precise and detailed prompt for each local patch.

\begin{figure}[t]
    \centering
    \includegraphics[width=1.0\linewidth]{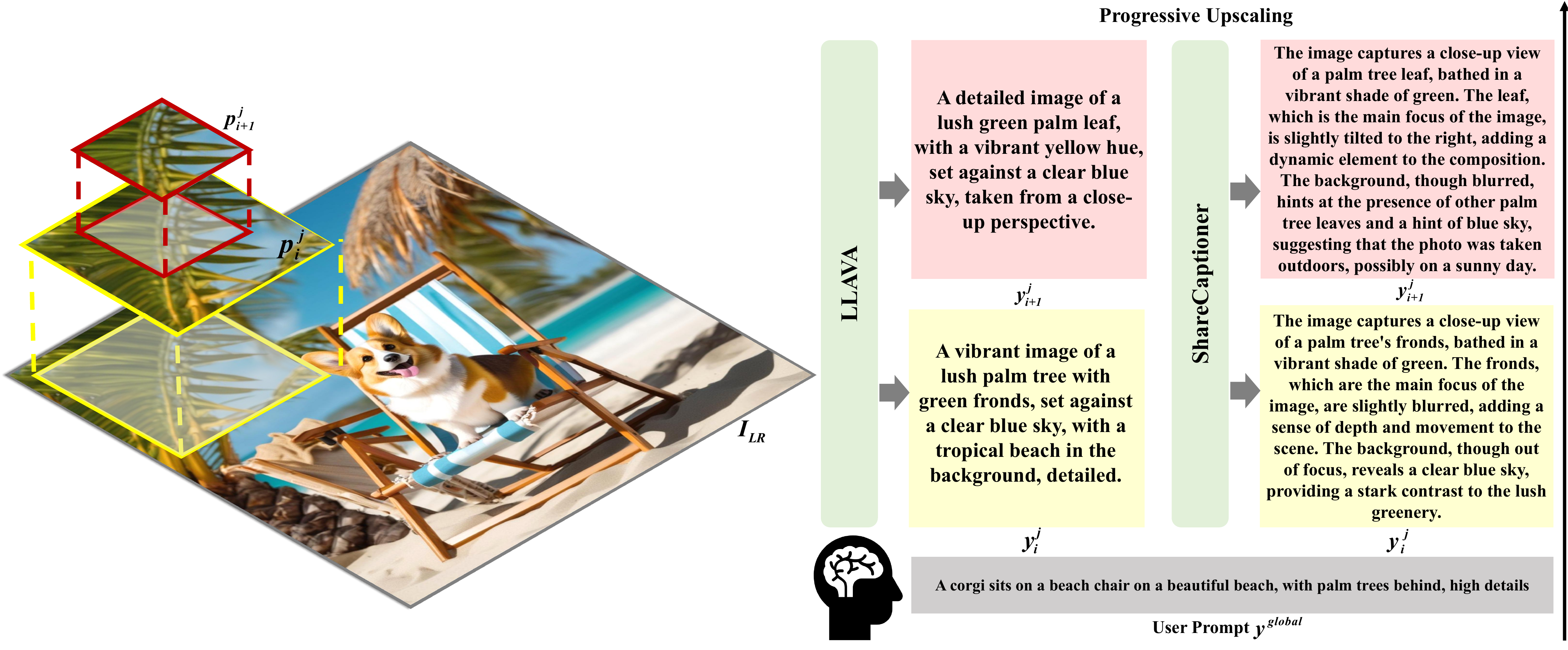}
    \vspace{-3mm}
    \caption{Comparion of hierarchical prompts from LLAVA~\cite{liu2024improved} and ShareCaptioner~\cite{chen2023sharegpt4v}. MLLM generates a dense local prompt $\mathbf{y}_{i}^j$ to describe details and textures of each local patch $\mathbf{p}_{i}^j$.}
    \label{fig:mllms}
\end{figure}

As illustrated in Fig.~\ref{fig:mllms}, the prompt generated by LLAVA precisely identifies background elements such as the ``palm tree'' and ``blue sky'', while appropriately excluding references to objects like ``corgi dog'' or ``beach chair'', which are mentioned in the global prompt but absent from the image patch. 
Compared to LLAVA, ShareCaptioner is capable of providing more detailed descriptions of relative object positions and visual focal points, which contributes to the generation of higher-quality images.




\noindent\textbf{N-grams Refinement.}
Occasionally, when local patches are unrecognizable, pre-trained MLLMs produce descriptions that are irrelevant to the global image. To mitigate the introduction of noise by the MLLMs, we generate text queries using N-grams $(n = 1)$ derived from local patch prompts and exclude unrelated tokens based on the similarity scores between image patches and dense descriptions. Before matching N-grams with patches, we filter out N-grams that do not form informative or grammatically correct captions for local patches. 
This process involves two stages: first, removing captions composed solely of uninformative words (e.g., \textit{image}, \textit{jpg}, \textit{background}), and second, eliminating articles (e.g., \textit{a}, \textit{the}) and prepositions (e.g., \textit{to}, \textit{of}, \textit{on}, \textit{in}) from the N-grams.
Subsequently, we exclude irrelevant tokens by filtering out those with scores below the average text-to-patch similarity threshold to avoid introducing information that was not present in the original user prompt.

Following the refined instructions, HiPrompt produces more accurate structures and richer local descriptions, ultimately enhancing visual quality.
\subsection{Noisy Image Decomposition}
Intending to produce consistent semantics and structures across various scales, we conduct a noisy image decomposition to achieve simultaneous denoising. The high and low spatial frequency components are controlled by local and global descriptions during the image generation process.
As illustrated in the lower part of Figure~\ref{fig:architecture}.
during the inverse denoising process, we utilize hierarchical prompts to estimate noise, which is then aggregated to denoise the image.
In our hierarchical prompt-based diffusion pipeline, we explore a decomposition of the noisy image \( \mathbf{z}_t \in \mathbb{R}^{ H \times W \times 3 } \) into two components: \( v_{\text{low}}(\mathbf{z}_t) \) and \( v_{\text{high}}(\mathbf{z}_t) \).

\begin{equation}
\mathbf{z}_t=\underbrace{\mathbf{z}_t-G_\sigma(\mathbf{z}_t)}_{v_{\text {high}}(\mathbf{z}_t)}+\underbrace{G_\sigma(\mathbf{z}_t)}_{v_{\text {low}}(\mathbf{z}_t)},
\end{equation}
where $G_\sigma$is a low-pass filter implemented as a Gaussian blur with standard
deviation $\sigma$, and $\mathbf{z}_t-G_\sigma(\mathbf{z}_t)$ acts as a high pass of $\mathbf{z}_t$. A higher standard
deviation $\sigma$ corresponds to a lower cut-off frequency on the high-pass filter, thereby making the high-pass prompts more prominent in the results. In our experimental setup, $\sigma$  is set to 2. 
Then, we can align the high-frequency component with a revised, accurate, and dense caption $\mathbf{y}^k$, while the low-frequency component is matched with the global user description.

\noindent\textbf{Parallel Denoising.} 
During the inverse denoising process, hierarchical prompts are employed to denoise multiple conditions of an image simultaneously. Specifically, two distinct prompts are used, each associated with a spatial filter function \( v_k(\cdot) \) that applies low and high pass filters to a noisy image. Given a diffusion model \( \epsilon_\theta(\cdot) \) and a partially denoised image \( \mathbf{z}_t \), noise estimated from different prompts are combined into a single estimate by summation as

\begin{equation}
\tilde{\epsilon}_t = \sum_k \epsilon_\theta\left(\mathbf{z}_t, \mathbf{y}^k, t\right).
\end{equation}

Here, \( \epsilon_k = \epsilon_\theta(\mathbf{z}_t, \mathbf{y}^k, t) \) represents the estimated noise conditioned on each hierarchical prompt \( \mathbf{y}^k \). Each filter transforms the noisy image \( \mathbf{z}_t \), providing noise estimates for the transformed images. These noise estimates are then aggregated to produce a combined estimate, which is subsequently applied within the diffusion sampling process.
Our method avoids transitioning to the frequency domain and instead controls directly in the spatial domain, which simplifies computations and offers a more direct approach. Additionally, we have designed a noise decomposition module that enhances structural consistency across resolutions from both high- and low-frequency perspectives
\section{Experiments}
In this section, we report qualitative and quantitative experiments and ablation studies. We validate the performance of HiPrompt based on the SDXL~\cite{podell2023sdxlimprovinglatentdiffusion}.
\begin{table*}[t]
\centering
  \setlength\tabcolsep{2pt}
  \centering
  \begin{tabular}{@{}clccccccc@{}}
    \toprule 
    Resolution & Method & $\text{FID}_r\downarrow $ & $\text{KID}_r\downarrow$   & $\text{FID}_c\downarrow$& $\text{KID}_c\downarrow$& $\text{IS}_r\uparrow$  & $\text{CLIP}\uparrow$& Time\\
    \midrule
    1024 $\times$ 1024 (1$\times$)    & SDXL-DI~\cite{podell2023sdxl}  &68.48  &0.0031  &69.23&0.0346&21.29&32.58& $<1$ min\\
    \midrule
    \multirow{6}{*}{2048 $\times$ 2048 (4$\times$)} & SDXL-DI~\cite{podell2023sdxl} &122.12 &0.0267&71.51&0.0371&12.50&29.27& 1 min\\
                                                    & MultiDiffusion~\cite{bar2023multidiffusion} & 139.12 & 0.0345& 86.73 &0.0483&12.34&29.27& 2 min\\
                                                    & ScaleCrafter~\cite{he2023scalecrafter} & 83.51 & 0.0074  & 57.29 &0.0225&15.36&30.50& 1 min\\ 
                                                    & DemoFusion~\cite{du2024demofusion}  & \underline{68.16} & \underline{0.0043}  & \underline{37.30} &\underline{0.0175}&\underline{19.24}&\textbf{32.62}& 2 min\\ 
                                                    & Ours &\textbf{67.79} &\textbf{0.0038}&\textbf{35.82}&\textbf{0.0170}&\textbf{19.37}&\underline{32.51}& 2 min\\
    \midrule
    \multirow{5}{*}{2048 $\times$ 4096 (8$\times$) }  & SDXL-DI~\cite{podell2023sdxl} &211.02 &0.0917&89.46&0.0428&7.96&25.77& 3 min\\
                                                    & MultiDiffusion~\cite{bar2023multidiffusion} & 212.58 & 0.0887& 86.14&0.0475&8.26&28.35& 3 min\\ 
                                                    & ScaleCrafter~\cite{he2023scalecrafter}  & 116.60 &0.0238 &72.91&0.0266  &11.06&26.32& 3 min\\  
                                                    & DemoFusion~\cite{du2024demofusion}  & \underline{76.25} & \underline{0.0076} &\underline{40.67} &\textbf{0.0093}&\textbf{16.68}&\underline{29.97}& 5 min\\ 
                                                    & Ours &\textbf{73.82} &\textbf{0.0076}&\textbf{35.17}&\underline{0.0122}&\underline{16.46}&\textbf{30.58}&5 min \\
    \midrule
    \multirow{6}{*}{4096 $\times$ 4096 (16$\times$)}  & SDXL-DI~\cite{podell2023sdxl} &231.22 &0.0935&90.36&0.0470&7.59&23.41& 5 min\\
    
                                                    & MultiDiffusion~\cite{bar2023multidiffusion} &269.65  &0.1305     &87.82&0.0486&7.36&23.98& 10 min\\
                                                    & ScaleCrafter~\cite{he2023scalecrafter} &109.49  & 0.0186  & 63.11 &0.0197&12.15&28.05& 9 min\\     
                                                    & DemoFusion~\cite{du2024demofusion}  & \underline{71.04} & \underline{0.0058} &  \textbf{36.89}&\underline{0.0178}&\underline{19.19}&\textbf{32.46}& 6 min\\ 
                                                    & Ours& \textbf{70.40}&\textbf{0.0053}&\underline{43.60}&\textbf{0.0162}&\textbf{19.20}&\underline{31.93}&8 min \\
  \bottomrule
  \end{tabular}
  \caption{Comparison of quantitative metrics between different training-free image generation extrapolation methods. We mark the best results in bold and underline to emphasize the second-best result. 
  }
  \label{tab:quantitative comparison}
\end{table*}
\subsection{Experimental Setup}
We compare HiPrompt with the following competitive approaches: (i) SDXL Direct Inference,
(ii) MultiDiffusion~\cite{bar2023multidiffusion},
(iii) ScaleCrafter~\cite{he2023scalecrafter},
and (iv) DemoFusion~\cite{du2024demofusion}.
We comprehensively evaluate the performance of our model at resolutions of $2048^2,2048 \times 4096$, and $4096^2$. Additionally, we leverage LLAVA-V1.6 and InternLM-Xcomposer-7B based ShareCaptioner~\cite{chen2023sharegpt4v} to generate hierarchical prompts of low-resolution image patches. 

\subsection{Quantitative Results}
To fairly evaluate the performance of the models, we perform quantitative experiments on the dataset of Laion-5B~\cite{schuhmann2022laion5bopenlargescaledataset} with a large number of image-caption pairs. We randomly sample 1k captions as the text prompts for the high-resolution image generation. We adopt four metrics following prior works~\cite{du2024demofusion}: Frechet Inception Distance (FID)~\cite{heusel2018ganstrainedtimescaleupdate}, Kernel Inception Distance (KID)~\cite{bińkowski2021demystifyingmmdgans}, Inception Score (IS)~\cite{salimans2016improvedtechniquestraininggans} and CLIP Score~\cite{radford2021learning} to evaluate both image quality and semantic similarity between image features and text prompts.
$\text{FID}_r$,$\text{KID}_r$ and $\text{IS}_r$ are used to measure the overall generation performance. Among them, FID and KID require resizing the test images to $299^2$, which may influence the evaluation results for higher-resolution images. For more reasonable evaluation, we follow CLEAN-FID~\cite{parmar2022aliased} to crop and resize some local patches 
to compute FID and KID, referred to as $\text{FID}_c$/$\text{KID}_c$.
We report quantitative results at three different resolutions. The inference time consumption is measured on a single NVIDIA H800 GPU.

As shown in Table~\ref{tab:quantitative comparison}, HiPrompt achieves significant performance gain over existing methods across most of the metrics. Specifically, HiPrompt outperforms prior state-of-the-art work~\cite{du2024demofusion} by 2.43 and 5.5 on $\text{FID}_r$ and $\text{FID}_c$ metrics at resolution of $2048 \times 4096$. By leveraging hierarchical prompts and an innovative noisy image decomposition design, HiPrompt effectively mitigates object repetition and synthesizes more precise textures and details. Given that the existing metrics do not fully capture the repetition issue, we present visual comparisons in Figure~\ref{fig:quality} to further illustrate this aspect. 

\subsection{Qualitative Results}
\begin{figure}[h]
    \centering
    \includegraphics[height=1.5\linewidth]{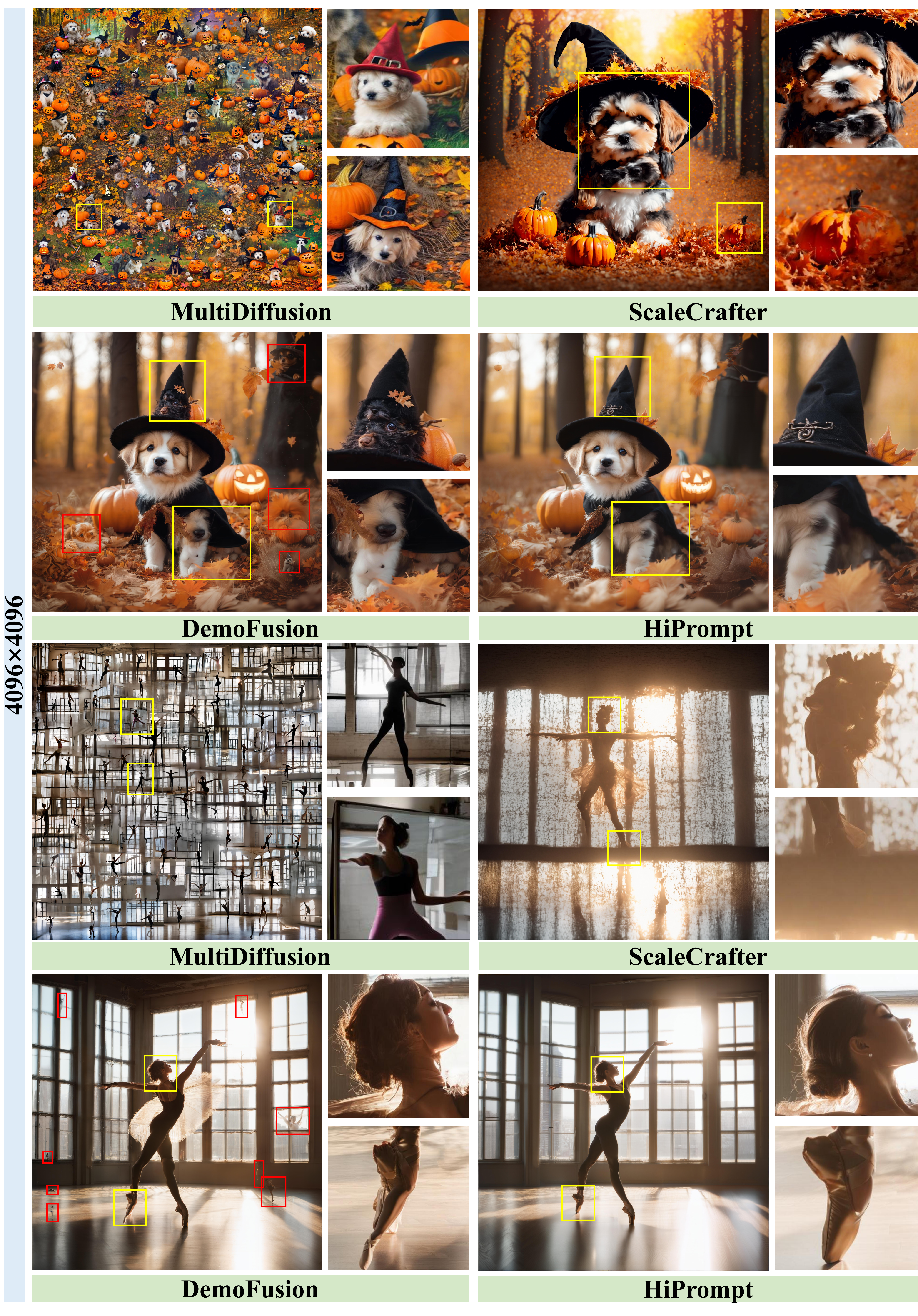}
    \vspace{-3mm}
    \caption{Qualitative comparison with other baselines. The red boxes highlight the repeated small objects, and the yellow boxes denote blurred areas and unreasonable structures.}
    \label{fig:quality}
\end{figure}
        Figure~\ref{fig:quality} illustrates a visual comparison between HiPrompt and other tuning-free higher-resolution image synthesis approaches. Each model produces outputs at a 16$\times$ resolution ($4096^2$) for comparison with the original SDXL. 
        In the first scenario, HiPrompt excels in generating superior semantic coherence and fine-grained details without any repetitions even in examples susceptible to pattern confusion. Contrastly, MultiDiffusion suffers from severely repeated and distorted generation. ScaleCrafter produces visually unpleasant structures and large areas of irregular textures, significantly degrading visual quality. Also, DemoFusion appears many small-dog repetitions and unreasonable structure artifacts due to its insufficient patch-based generation and lack of fine-grained guidance for local content. 

Likewise, the results from the ballet dancer case further validate our observations. HiPrompt effectively restores the girl's clear facial features and refines the intricate structure of the ballet shoes, making them more precise, cohesive, and visually appealing in complex real-world scenarios. In opposition, ScaleCrafter exhibits weak structural preservation and chaotic details, while DemoFusion not only introduces multiple repetitive elements in the background but also creates unreasonable duplications of ballet shoes distorting the human figure.

\subsection{Ablation Studies}
In this section, we first perform ablation studies on the two core modules of HiPrompt and then discuss the effects of different MLLMs and N-grams refinement modules.

 \begin{table}[t]
\centering
\tabcolsep=0.35cm
\begin{tabular}{cc|ccc}
    \toprule
    MLLM & ND & FID$_r$ $\downarrow$ & FID$_c$ $\downarrow$  & CLIP $\uparrow$ \\
    \midrule 
    $\mathbf{\times}$ & $\mathbf{\times}$  & 76.24  &40.67& 29.97  \\
    $\mathbf{\checkmark}$ & $\mathbf{\times}$ & 75.03 &46.93& 28.55 \\
    $\mathbf{\checkmark}$ & $\mathbf{\checkmark}$ & \textbf{73.82} &\textbf{35.17}& \textbf{30.58}  \\
    \bottomrule
\end{tabular}
\caption{Ablation study results of core components: Hierarchical MLLM Prompts Guidance, Noise Decomposition (ND). The best results are marked in bold. } 
\label{tab:fact_ablation}
\end{table}

\begin{table}[t]
\centering
\tabcolsep=0.15cm
\begin{tabular}{cc|c|cccc}
    \toprule
    LLAVA& SC& ND & FID$_r$ $\downarrow$ & KID$_r$ $\downarrow$ & FID$_c$ $\downarrow$ & KID$_c$ $\downarrow$  \\
    \midrule
    $\mathbf{\times}$&$\mathbf{\times}$&$\mathbf{\times}$  & 71.04 &0.0058 & 14.32&0.0059  \\
    $\mathbf{\checkmark}$&$\mathbf{\times}$&$\mathbf{\times}$  & 71.81 & 0.0062 &11.82 &0.0039\\
    $\mathbf{\times}$&$\mathbf{\checkmark}$&$\mathbf{\times}$    & 71.01&0.0060 & \textbf{8.80} & \textbf{0.0022}  \\
    $\mathbf{\checkmark}$&$\mathbf{\times}$&$\mathbf{\checkmark}$  & 70.40 & 0.0053 & 14.01 &0.0059 \\
    $\mathbf{\times}$&$\mathbf{\checkmark}$&$\mathbf{\checkmark}$    & \textbf{70.22} & \textbf{0.0051} & 12.16&0.0045  \\
    \bottomrule
\end{tabular}
\caption{Ablation study of different MLLMs including LLAVA~\cite{liu2024improved} and ShareCaptioner (SC)~\cite{chen2023sharegpt4v}. The best results are marked in bold.} 
\label{tab: mllm_ablation}
\end{table}

\begin{figure}[h]
    \centering
    \includegraphics[width=1.\linewidth]{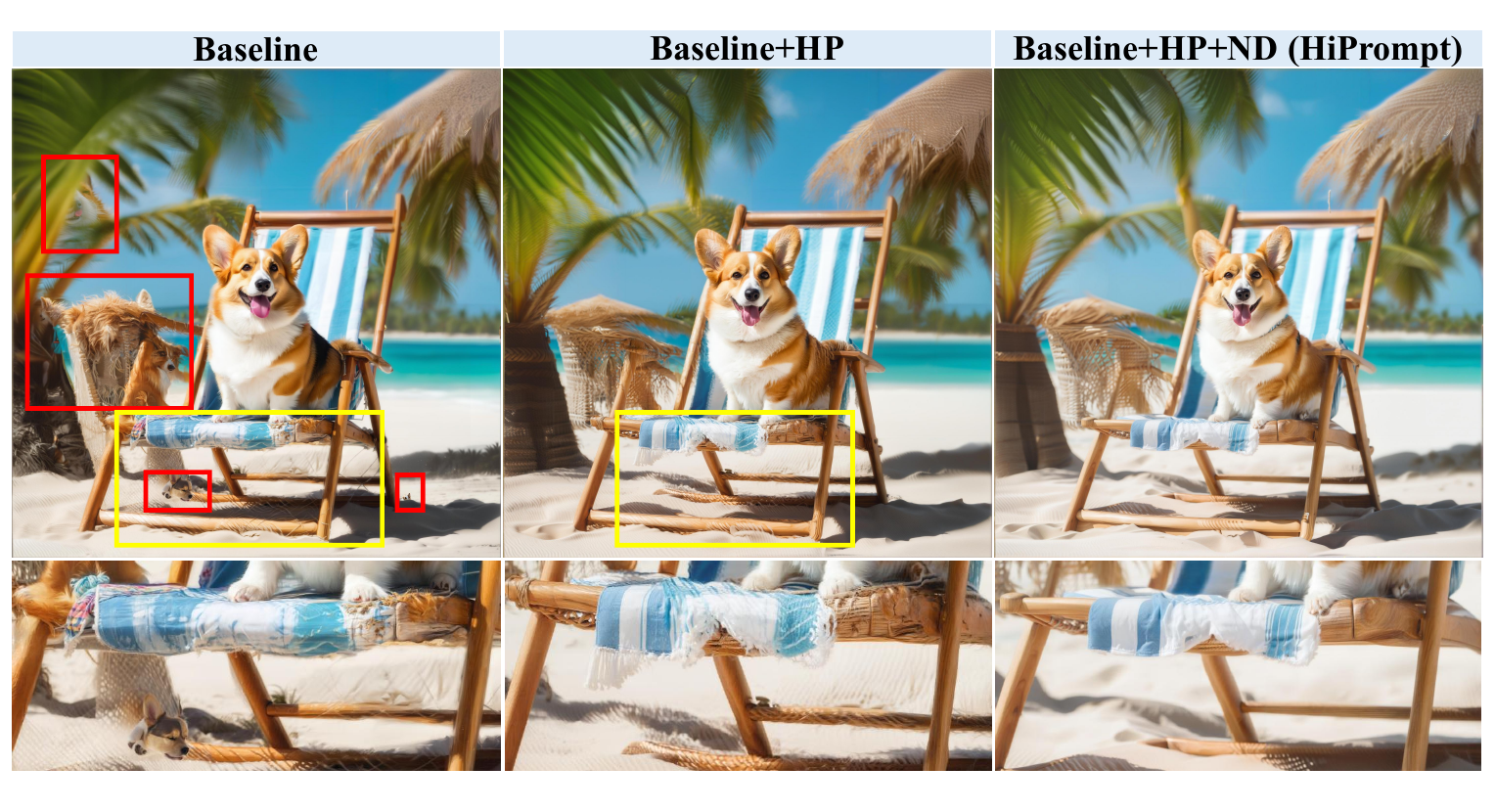}
    \caption{The ablation study of three components used in our approach: Hierarchical MLLM Prompts guidance (HP), Noise Decomposition (ND). All results are presented at a resolution of $4096^2$ (16$\times$). The second row presents the local details that have been zoomed in.}
    \vspace{-3mm}
    \label{fig:ablation_fact}
\end{figure}

\noindent\textbf{Effects of Core Components.} We conduct ablation studies on the two components of HiPrompt: hierarchical MLLM prompts guidance and noise decomposition. As depicted in Figure~\ref{fig:ablation_fact}, the absence of any module leads to a decline in generation quality. We present the result of the baseline model ~\cite{du2024demofusion}, exhibiting structural distortions and repeated patterns. Without hierarchical patch-wise prompts, the resulting image contains numerous repeated small objects, emphasizing the importance of patch-content-aware prompts in preventing the generation of repetitive elements. Furthermore, when we introduce the noise decomposition strategy to parallel denoising, the issue of structural distortions is resolved.
 This implies that the two modules work together to effectively alleviate repetitive objects and enhance the image quality. In conclusion, when combined, they leverage their respective strengths and functionalities, resulting in impressive generative outcomes. The quantitative results of the core components ablation study are shown in  Table~\ref{tab:fact_ablation} and all the results are performed in the resolution of $2048 \times 4096$ with LLAVA~\cite{liu2024improved}. Compared to the baseline (first row), adding the hierarchical MLLM prompt guidance results in a 1.21 decrease in FID$_r$. Furthermore, if both the hierarchical MLLM prompt guidance and Noise Decomposition (ND) modules are incorporated, HiPrompt surpasses the baseline on three metrics, and the FID$_r$ can be decreased by 2.42 ($76.24 \longrightarrow 73.82$).

\noindent\textbf{Effects of Different MLLMs.}
\begin{table}[t]
\centering
\tabcolsep=0.45cm
\begin{tabular}{cc|ccc}
    \toprule
    MLLM & NR & FID$_r$ $\downarrow$ & KID$_r$ $\downarrow$& $\text{IS}_r\uparrow$  \\
    \midrule
    $\mathbf{\checkmark}$ & $\mathbf{\times}$    & 68.78 &0.0043&19.29\\
    $\mathbf{\checkmark}$ & $\mathbf{\checkmark}$      & \textbf{67.79}&\textbf{0.0038}&\textbf{19.37}  \\
    \bottomrule
\end{tabular}
\caption{Ablation study of N-grams Refinement (NR).The best results are marked in bold.} 
\label{tab: ngrams_study}
\end{table}
We investigate the use of various MLLMs, such as LLAVA~\cite{liu2024improved} and ShareCaptioner~\cite{chen2023sharegpt4v}, to validate the effectiveness of our hierarchical prompt-based pipeline. These models generate corresponding dense and accurate local prompts to reduce the repetitions. As shown in Table~\ref{tab: mllm_ablation}, we conduct ablation experiments at a resolution of $4096^2$, which demonstrate that both LLAVA and ShareCaptioner significantly enhance the quality of high-resolution image generation. In particular, introducing them both lead to decreases of 2.5 and 5.52 in FID$_c$ respectively, and the effect of ShareCaptioner is better than that of LLAVA because ShareCaptioner itself generates more dense descriptions. Additionally, the HiPrompt is shown to be universally compatible with mainstream MLLMs.

\noindent\textbf{Effects of N-grams Refinement.}
We then assess the impact of the proposed N-grams refinement strategy at a resolution of $2048^2$ on image generation. As demonstrated in Table~\ref{tab: ngrams_study}, the first row presents HiPrompt combined with LLAVA~\cite{liu2024improved}, while the second row shows HiPrompt combined with LLAVA and N-grams refinement. The addition of n-gram refinement significantly improves the image quality, further demonstrating the importance of accurate captions for higher-resolution image generation. 

\section{Conclusion}
In this paper, we propose HiPrompt, an effective framework of tuning-free higher-resolution image generation with hierarchical prompts from MLLMs.
To address the object repetition and structure distortion issues, we first introduce a hierarchical prompts-based diffusion model that utilizes patch-wise dense descriptions from MLLMs to elaborately guide the local structure and texture generation to avoid pattern repetition radically. To parallel denoise conditioning on hierarchical prompts during the inverse denoising process, we decompose the generated noisy image into low and high spatial frequencies. Then summarize both the estimated predictions to the final results which aligns with hierarchical prompt-based guidance. Additionally, we explore different MLLMs and empirically validate that they could achieve noticeable enhancement. Extensive qualitative and quantitative experiments demonstrate the validity of our approach in conducting higher-resolution image generation of high quality.

\bibliography{aaai25}

\begin{thebibliography}{44}
\providecommand{\natexlab}[1]{#1}

\bibitem[{Bar-Tal et~al.(2023)Bar-Tal, Yariv, Lipman, and Dekel}]{bar2023multidiffusion}
Bar-Tal, O.; Yariv, L.; Lipman, Y.; and Dekel, T. 2023.
\newblock Multidiffusion: Fusing diffusion paths for controlled image generation.

\bibitem[{Bińkowski et~al.(2021)Bińkowski, Sutherland, Arbel, and Gretton}]{bińkowski2021demystifyingmmdgans}
Bińkowski, M.; Sutherland, D.~J.; Arbel, M.; and Gretton, A. 2021.
\newblock Demystifying MMD GANs.
\newblock arXiv:1801.01401.

\bibitem[{Chen et~al.(2023{\natexlab{a}})Chen, Yu, Ge, Yao, Xie, Wu, Wang, Kwok, Luo, Lu et~al.}]{chen2023pixart}
Chen, J.; Yu, J.; Ge, C.; Yao, L.; Xie, E.; Wu, Y.; Wang, Z.; Kwok, J.; Luo, P.; Lu, H.; et~al. 2023{\natexlab{a}}.
\newblock Pixart-$\alpha$: Fast training of diffusion transformer for photorealistic text-to-image synthesis.
\newblock \emph{arXiv preprint arXiv:2310.00426}.

\bibitem[{Chen et~al.(2023{\natexlab{b}})Chen, Li, Dong, Zhang, He, Wang, Zhao, and Lin}]{chen2023sharegpt4v}
Chen, L.; Li, J.; Dong, X.; Zhang, P.; He, C.; Wang, J.; Zhao, F.; and Lin, D. 2023{\natexlab{b}}.
\newblock Sharegpt4v: Improving large multi-modal models with better captions.
\newblock \emph{arXiv preprint arXiv:2311.12793}.

\bibitem[{Chen, Laina, and Vedaldi(2024)}]{chen2024training}
Chen, M.; Laina, I.; and Vedaldi, A. 2024.
\newblock Training-free layout control with cross-attention guidance.
\newblock In \emph{Proceedings of the IEEE/CVF Winter Conference on Applications of Computer Vision}, 5343--5353.

\bibitem[{Chen et~al.(2024)Chen, Wang, Zhang, Shechtman, Wang, and Gharbi}]{chen2024image}
Chen, Y.; Wang, O.; Zhang, R.; Shechtman, E.; Wang, X.; and Gharbi, M. 2024.
\newblock Image Neural Field Diffusion Models.
\newblock In \emph{Proceedings of the IEEE/CVF Conference on Computer Vision and Pattern Recognition}, 8007--8017.

\bibitem[{Ding et~al.(2021)Ding, Yang, Hong, Zheng, Zhou, Yin, Lin, Zou, Shao, Yang et~al.}]{ding2021cogview}
Ding, M.; Yang, Z.; Hong, W.; Zheng, W.; Zhou, C.; Yin, D.; Lin, J.; Zou, X.; Shao, Z.; Yang, H.; et~al. 2021.
\newblock Cogview: Mastering text-to-image generation via transformers.
\newblock \emph{Advances in neural information processing systems}, 34: 19822--19835.

\bibitem[{Ding et~al.(2023)Ding, Zhang, Wu, and Tu}]{ding2023patched}
Ding, Z.; Zhang, M.; Wu, J.; and Tu, Z. 2023.
\newblock Patched denoising diffusion models for high-resolution image synthesis.
\newblock In \emph{The Twelfth International Conference on Learning Representations}.

\bibitem[{Du et~al.(2024)Du, Chang, Hospedales, Song, and Ma}]{du2024demofusion}
Du, R.; Chang, D.; Hospedales, T.; Song, Y.-Z.; and Ma, Z. 2024.
\newblock Demofusion: Democratising high-resolution image generation with no \$\$\$.
\newblock In \emph{Proceedings of the IEEE/CVF Conference on Computer Vision and Pattern Recognition}, 6159--6168.

\bibitem[{Feng et~al.(2024)Feng, Gong, Chen, Shen, Liu, and Zhou}]{feng2024ranni}
Feng, Y.; Gong, B.; Chen, D.; Shen, Y.; Liu, Y.; and Zhou, J. 2024.
\newblock Ranni: Taming text-to-image diffusion for accurate instruction following.
\newblock In \emph{Proceedings of the IEEE/CVF Conference on Computer Vision and Pattern Recognition}, 4744--4753.

\bibitem[{Geng, Park, and Owens(2024)}]{geng2024visual}
Geng, D.; Park, I.; and Owens, A. 2024.
\newblock Visual anagrams: Generating multi-view optical illusions with diffusion models.
\newblock In \emph{Proceedings of the IEEE/CVF Conference on Computer Vision and Pattern Recognition}, 24154--24163.

\bibitem[{Gu et~al.(2023)Gu, Zhai, Zhang, Susskind, and Jaitly}]{gu2023matryoshka}
Gu, J.; Zhai, S.; Zhang, Y.; Susskind, J.~M.; and Jaitly, N. 2023.
\newblock Matryoshka diffusion models.
\newblock In \emph{The Twelfth International Conference on Learning Representations}.

\bibitem[{Guo et~al.(2024)Guo, He, Chen, Xia, Cun, Wang, Huang, Zhang, Wang, Chen et~al.}]{guo2024make}
Guo, L.; He, Y.; Chen, H.; Xia, M.; Cun, X.; Wang, Y.; Huang, S.; Zhang, Y.; Wang, X.; Chen, Q.; et~al. 2024.
\newblock Make a cheap scaling: A self-cascade diffusion model for higher-resolution adaptation.
\newblock \emph{arXiv preprint arXiv:2402.10491}.

\bibitem[{He et~al.(2023)He, Yang, Chen, Cun, Xia, Zhang, Wang, He, Chen, and Shan}]{he2023scalecrafter}
He, Y.; Yang, S.; Chen, H.; Cun, X.; Xia, M.; Zhang, Y.; Wang, X.; He, R.; Chen, Q.; and Shan, Y. 2023.
\newblock Scalecrafter: Tuning-free higher-resolution visual generation with diffusion models.
\newblock In \emph{The Twelfth International Conference on Learning Representations}.

\bibitem[{Heusel et~al.(2018)Heusel, Ramsauer, Unterthiner, Nessler, and Hochreiter}]{heusel2018ganstrainedtimescaleupdate}
Heusel, M.; Ramsauer, H.; Unterthiner, T.; Nessler, B.; and Hochreiter, S. 2018.
\newblock GANs Trained by a Two Time-Scale Update Rule Converge to a Local Nash Equilibrium.
\newblock arXiv:1706.08500.

\bibitem[{Ho, Jain, and Abbeel(2020)}]{ho2020denoising}
Ho, J.; Jain, A.; and Abbeel, P. 2020.
\newblock Denoising diffusion probabilistic models.
\newblock \emph{Advances in neural information processing systems}, 33: 6840--6851.

\bibitem[{Huang et~al.(2024)Huang, Fang, Zhang, Song, Liu, Liu, and Li}]{huang2024fouriscale}
Huang, L.; Fang, R.; Zhang, A.; Song, G.; Liu, S.; Liu, Y.; and Li, H. 2024.
\newblock FouriScale: A Frequency Perspective on Training-Free High-Resolution Image Synthesis.
\newblock \emph{arXiv preprint arXiv:2403.12963}.

\bibitem[{Jin et~al.(2024)Jin, Shen, Li, and Xue}]{jin2024training}
Jin, Z.; Shen, X.; Li, B.; and Xue, X. 2024.
\newblock Training-free diffusion model adaptation for variable-sized text-to-image synthesis.
\newblock \emph{Advances in Neural Information Processing Systems}, 36.

\bibitem[{Li et~al.(2024)Li, Kamko, Akhgari, Sabet, Xu, and Doshi}]{li2024playground}
Li, D.; Kamko, A.; Akhgari, E.; Sabet, A.; Xu, L.; and Doshi, S. 2024.
\newblock Playground v2. 5: Three insights towards enhancing aesthetic quality in text-to-image generation.
\newblock \emph{arXiv preprint arXiv:2402.17245}.

\bibitem[{Lin et~al.(2024)Lin, Lin, Zhao, and Ji}]{lin2024accdiffusion}
Lin, Z.; Lin, M.; Zhao, M.; and Ji, R. 2024.
\newblock AccDiffusion: An Accurate Method for Higher-Resolution Image Generation.
\newblock \emph{arXiv preprint arXiv:2407.10738}.

\bibitem[{Liu et~al.(2024)Liu, Li, Li, and Lee}]{liu2024improved}
Liu, H.; Li, C.; Li, Y.; and Lee, Y.~J. 2024.
\newblock Improved baselines with visual instruction tuning.
\newblock In \emph{Proceedings of the IEEE/CVF Conference on Computer Vision and Pattern Recognition}, 26296--26306.

\bibitem[{Lu et~al.(2024)Lu, Wang, Huang, Wu, Liu, Ouyang, and Bai}]{lu2024fit}
Lu, Z.; Wang, Z.; Huang, D.; Wu, C.; Liu, X.; Ouyang, W.; and Bai, L. 2024.
\newblock Fit: Flexible vision transformer for diffusion model.
\newblock \emph{arXiv preprint arXiv:2402.12376}.

\bibitem[{Parmar, Zhang, and Zhu(2022)}]{parmar2022aliased}
Parmar, G.; Zhang, R.; and Zhu, J.-Y. 2022.
\newblock On aliased resizing and surprising subtleties in gan evaluation.
\newblock In \emph{Proceedings of the IEEE/CVF Conference on Computer Vision and Pattern Recognition}, 11410--11420.

\bibitem[{Podell et~al.(2023{\natexlab{a}})Podell, English, Lacey, Blattmann, Dockhorn, M{\"u}ller, Penna, and Rombach}]{podell2023sdxl}
Podell, D.; English, Z.; Lacey, K.; Blattmann, A.; Dockhorn, T.; M{\"u}ller, J.; Penna, J.; and Rombach, R. 2023{\natexlab{a}}.
\newblock Sdxl: Improving latent diffusion models for high-resolution image synthesis.
\newblock \emph{arXiv preprint arXiv:2307.01952}.

\bibitem[{Podell et~al.(2023{\natexlab{b}})Podell, English, Lacey, Blattmann, Dockhorn, Müller, Penna, and Rombach}]{podell2023sdxlimprovinglatentdiffusion}
Podell, D.; English, Z.; Lacey, K.; Blattmann, A.; Dockhorn, T.; Müller, J.; Penna, J.; and Rombach, R. 2023{\natexlab{b}}.
\newblock SDXL: Improving Latent Diffusion Models for High-Resolution Image Synthesis.
\newblock arXiv:2307.01952.

\bibitem[{Radford et~al.(2021)Radford, Kim, Hallacy, Ramesh, Goh, Agarwal, Sastry, Askell, Mishkin, Clark et~al.}]{radford2021learning}
Radford, A.; Kim, J.~W.; Hallacy, C.; Ramesh, A.; Goh, G.; Agarwal, S.; Sastry, G.; Askell, A.; Mishkin, P.; Clark, J.; et~al. 2021.
\newblock Learning transferable visual models from natural language supervision.
\newblock In \emph{International conference on machine learning}, 8748--8763. PMLR.

\bibitem[{Rombach et~al.(2022{\natexlab{a}})Rombach, Blattmann, Lorenz, Esser, and Ommer}]{rombach2022highresolutionimagesynthesislatent}
Rombach, R.; Blattmann, A.; Lorenz, D.; Esser, P.; and Ommer, B. 2022{\natexlab{a}}.
\newblock High-Resolution Image Synthesis with Latent Diffusion Models.
\newblock arXiv:2112.10752.

\bibitem[{Rombach et~al.(2022{\natexlab{b}})Rombach, Blattmann, Lorenz, Esser, and Ommer}]{rombach2022high}
Rombach, R.; Blattmann, A.; Lorenz, D.; Esser, P.; and Ommer, B. 2022{\natexlab{b}}.
\newblock High-resolution image synthesis with latent diffusion models.
\newblock In \emph{Proceedings of the IEEE/CVF conference on computer vision and pattern recognition}, 10684--10695.

\bibitem[{Ronneberger, Fischer, and Brox(2015)}]{ronneberger2015u}
Ronneberger, O.; Fischer, P.; and Brox, T. 2015.
\newblock U-net: Convolutional networks for biomedical image segmentation.
\newblock In \emph{Medical image computing and computer-assisted intervention--MICCAI 2015: 18th international conference, Munich, Germany, October 5-9, 2015, proceedings, part III 18}, 234--241. Springer.

\bibitem[{Saharia et~al.(2022)Saharia, Chan, Saxena, Li, Whang, Denton, Ghasemipour, Gontijo~Lopes, Karagol~Ayan, Salimans et~al.}]{saharia2022photorealistic}
Saharia, C.; Chan, W.; Saxena, S.; Li, L.; Whang, J.; Denton, E.~L.; Ghasemipour, K.; Gontijo~Lopes, R.; Karagol~Ayan, B.; Salimans, T.; et~al. 2022.
\newblock Photorealistic text-to-image diffusion models with deep language understanding.
\newblock \emph{Advances in neural information processing systems}, 35: 36479--36494.

\bibitem[{Salimans et~al.(2016)Salimans, Goodfellow, Zaremba, Cheung, Radford, and Chen}]{salimans2016improvedtechniquestraininggans}
Salimans, T.; Goodfellow, I.; Zaremba, W.; Cheung, V.; Radford, A.; and Chen, X. 2016.
\newblock Improved Techniques for Training GANs.
\newblock arXiv:1606.03498.

\bibitem[{Schuhmann et~al.(2022)Schuhmann, Beaumont, Vencu, Gordon, Wightman, Cherti, Coombes, Katta, Mullis, Wortsman, Schramowski, Kundurthy, Crowson, Schmidt, Kaczmarczyk, and Jitsev}]{schuhmann2022laion5bopenlargescaledataset}
Schuhmann, C.; Beaumont, R.; Vencu, R.; Gordon, C.; Wightman, R.; Cherti, M.; Coombes, T.; Katta, A.; Mullis, C.; Wortsman, M.; Schramowski, P.; Kundurthy, S.; Crowson, K.; Schmidt, L.; Kaczmarczyk, R.; and Jitsev, J. 2022.
\newblock LAION-5B: An open large-scale dataset for training next generation image-text models.
\newblock arXiv:2210.08402.

\bibitem[{Shi et~al.(2024)Shi, Li, Zhang, He, Gong, and Zheng}]{shi2024resmastermasteringhighresolutionimage}
Shi, S.; Li, W.; Zhang, Y.; He, J.; Gong, B.; and Zheng, Y. 2024.
\newblock ResMaster: Mastering High-Resolution Image Generation via Structural and Fine-Grained Guidance.
\newblock arXiv:2406.16476.

\bibitem[{Si et~al.(2024)Si, Huang, Jiang, and Liu}]{si2024freeu}
Si, C.; Huang, Z.; Jiang, Y.; and Liu, Z. 2024.
\newblock Freeu: Free lunch in diffusion u-net.
\newblock In \emph{Proceedings of the IEEE/CVF Conference on Computer Vision and Pattern Recognition}, 4733--4743.

\bibitem[{Song, Meng, and Ermon(2020)}]{song2020denoising}
Song, J.; Meng, C.; and Ermon, S. 2020.
\newblock Denoising diffusion implicit models.
\newblock \emph{arXiv preprint arXiv:2010.02502}.

\bibitem[{Teng et~al.(2023)Teng, Zheng, Ding, Hong, Wangni, Yang, and Tang}]{teng2023relay}
Teng, J.; Zheng, W.; Ding, M.; Hong, W.; Wangni, J.; Yang, Z.; and Tang, J. 2023.
\newblock Relay diffusion: Unifying diffusion process across resolutions for image synthesis.
\newblock \emph{arXiv preprint arXiv:2309.03350}.

\bibitem[{Wang et~al.(2024)Wang, Kontkanen, Curless, Seitz, Kemelmacher-Shlizerman, Mildenhall, Srinivasan, Verbin, and Holynski}]{wang2024generative}
Wang, X.; Kontkanen, J.; Curless, B.; Seitz, S.~M.; Kemelmacher-Shlizerman, I.; Mildenhall, B.; Srinivasan, P.; Verbin, D.; and Holynski, A. 2024.
\newblock Generative powers of ten.
\newblock In \emph{Proceedings of the IEEE/CVF Conference on Computer Vision and Pattern Recognition}, 7173--7182.

\bibitem[{Wang et~al.(2023)Wang, Chen, Ma, Zhou, Huang, Wang, Yang, He, Yu, Yang et~al.}]{wang2023lavie}
Wang, Y.; Chen, X.; Ma, X.; Zhou, S.; Huang, Z.; Wang, Y.; Yang, C.; He, Y.; Yu, J.; Yang, P.; et~al. 2023.
\newblock Lavie: High-quality video generation with cascaded latent diffusion models.
\newblock \emph{arXiv preprint arXiv:2309.15103}.

\bibitem[{Xie et~al.(2023)Xie, Yao, Shi, Liu, Zhou, Liu, Li, and Li}]{xie2023difffit}
Xie, E.; Yao, L.; Shi, H.; Liu, Z.; Zhou, D.; Liu, Z.; Li, J.; and Li, Z. 2023.
\newblock Difffit: Unlocking transferability of large diffusion models via simple parameter-efficient fine-tuning.
\newblock In \emph{Proceedings of the IEEE/CVF International Conference on Computer Vision}, 4230--4239.

\bibitem[{Yang et~al.(2024)Yang, Yu, Meng, Xu, Ermon, and Bin}]{yang2024mastering}
Yang, L.; Yu, Z.; Meng, C.; Xu, M.; Ermon, S.; and Bin, C. 2024.
\newblock Mastering text-to-image diffusion: Recaptioning, planning, and generating with multimodal llms.
\newblock In \emph{Forty-first International Conference on Machine Learning}.

\bibitem[{Zhang et~al.(2023{\natexlab{a}})Zhang, Wu, Liu, Zhao, Ran, Gu, Gao, and Shou}]{zhang2023show}
Zhang, D.~J.; Wu, J.~Z.; Liu, J.-W.; Zhao, R.; Ran, L.; Gu, Y.; Gao, D.; and Shou, M.~Z. 2023{\natexlab{a}}.
\newblock Show-1: Marrying pixel and latent diffusion models for text-to-video generation.
\newblock \emph{arXiv preprint arXiv:2309.15818}.

\bibitem[{Zhang et~al.(2021)Zhang, Liang, Van~Gool, and Timofte}]{zhang2021designing}
Zhang, K.; Liang, J.; Van~Gool, L.; and Timofte, R. 2021.
\newblock Designing a practical degradation model for deep blind image super-resolution.
\newblock In \emph{Proceedings of the IEEE/CVF International Conference on Computer Vision}, 4791--4800.

\bibitem[{Zhang et~al.(2023{\natexlab{b}})Zhang, Chen, Zhao, Chen, Tang, Chen, Cao, and Liang}]{zhang2023hidiffusion}
Zhang, S.; Chen, Z.; Zhao, Z.; Chen, Z.; Tang, Y.; Chen, Y.; Cao, W.; and Liang, J. 2023{\natexlab{b}}.
\newblock HiDiffusion: Unlocking High-Resolution Creativity and Efficiency in Low-Resolution Trained Diffusion Models.
\newblock \emph{arXiv preprint arXiv:2311.17528}.

\bibitem[{Zheng et~al.(2024)Zheng, Guo, Deng, Han, Li, Xu, and Xu}]{zheng2024any}
Zheng, Q.; Guo, Y.; Deng, J.; Han, J.; Li, Y.; Xu, S.; and Xu, H. 2024.
\newblock Any-size-diffusion: Toward efficient text-driven synthesis for any-size hd images.
\newblock In \emph{Proceedings of the AAAI Conference on Artificial Intelligence}, volume~38, 7571--7578.

\end{thebibliography}

\end{document}